\documentclass[letterpaper, 10 pt, conference]{ieeeconf}
\IEEEoverridecommandlockouts
\usepackage{multirow}
\usepackage{cite}
\usepackage{comment}
\usepackage{amsmath,amssymb,amsfonts}
\usepackage{graphicx}
\usepackage{textcomp}
\usepackage{xcolor}
\usepackage{color}
\usepackage{algorithm}
\usepackage{algpseudocode}
\definecolor{blue}{rgb}{0, 0, 1}
\definecolor{red}{rgb}{1, 0, 0}
\definecolor{black}{rgb}{0, 0, 0}

\newcommand{\tuba}[1]{\textcolor{black}{#1}}

\newcommand{\yigit}[1]{\textcolor{purple}{#1}}
\usepackage{soul}

\title{Bidirectional Human Interactive AI Framework for Social Robot Navigation}

\author{Tuba Girgin$^{1,2,*}$ , Emre Girgin$^{1*}$ , Yigit Yildirim$^{2}$ , Emre Ugur$^{2}$ , Mehmet Haklidir$^{1}$\thanks{The project AIMS5.0 is supported by the Chips Joint Undertaking and its members, including the top-up funding by National Funding Authorities from involved countries under grant agreement no. 101112089.} \thanks{$^1$Robotics and Autonomous Systems Laboratory, TUBITAK BILGEM, 41470 Kocaeli, Turkey. $^2$Department of Computer Engineering, Bogazici University, Istanbul, Turkey. $^{*}$ Equal contribution. } }

\date{September 2023}

\begin{document}

\maketitle

\begin{abstract}
     
     Trustworthiness is a crucial concept in the context of human-robot interaction. Cooperative robots must be transparent regarding their decision-making process, especially when operating in a human-oriented environment. This paper presents a comprehensive end-to-end framework aimed at fostering trustworthy bidirectional human-robot interaction in collaborative environments for the social navigation of mobile robots.  In this framework, the robot communicates verbally while the human guides with gestures. Our method enables a mobile robot to predict the trajectory of people and adjust its route in a socially-aware manner. In case of conflict between human and robot decisions, detected through visual examination, the route is dynamically modified based on human preference while verbal communication is maintained. We present our pipeline, framework design, and preliminary experiments that form the foundation of our proposition.
     
\end{abstract}

\section{Introduction}

The growing demand for autonomy in industry necessitates increased reliability in Human-Robot Interaction \cite{sheridan2016human}. Considering that navigation is listed as a nonverbal interaction \cite{bartneck2020human}, enhancing the reliability of navigation is a desired capability for autonomous mobile robots. Due to the same reason, improving the explainability of AI-based solutions has also gained attention in recent studies \cite{samek2021explaining}. Although the intersection of both concepts yields a reliable and responsible AI in social navigation and HRI, which is a preferred attribute for industrial applications, the trustworthy interaction of humans and robots working in a collaborative environment is under-studied. 

While the concept of inter-human trust is quantitatively examined \cite{hancock2023and}, the trustworthiness of the AI algorithms is generally stressed by conducting an exhaustive search on the state space of the models \cite{brandao2021towards}, \cite{fox2017explainable}, segmenting plans for the monitoring purposes \cite{almagor2020explainable}, \cite{kottinger2021maps}, \cite{kottinger2022conflict}, and failure explanation \cite{das2021explainable}, \cite{diehl2023causal}. In these settings, a human expert is necessary for offline validation of decisions, while providing online explanations during operation for non-experts is an area that has received limited study.

Although robots that are aware of social dynamics in their environment adjust their behavior, they may unintentionally disrupt people if their intentions are not clearly expressed \cite{angelopoulos2022you}. Maintaining trust between intelligent robots and humans in a social environment is crucial for integrating novel approaches into ready-to-go systems since an increase in the comprehensibility of the robot's movements directly contributes to the comfort of the people \cite{mavrogiannis2019effects}. However, many studies in social robot navigation fail to consider bidirectional interaction and communication. This feature improves the potential applications of robots in human-centric environments by providing explanations for the actions taken by the robot to maintain long-term collaboration and trust with humans.

\begin{figure}
    \centering
    \includegraphics[ width=0.3\textwidth]{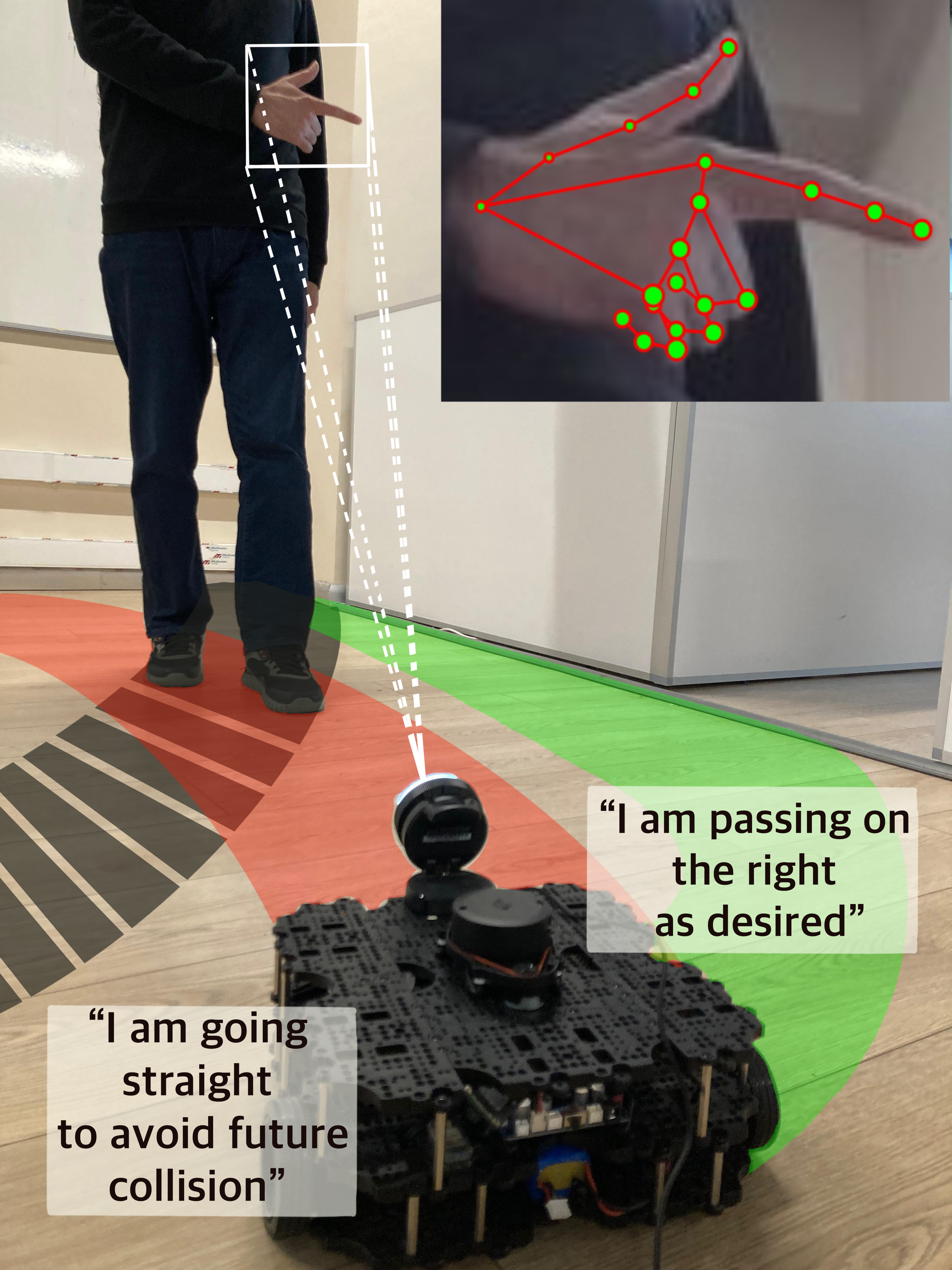}
    \caption{An example trustworthy social navigation scenario is illustrated. First, the robot plans a path (red) avoiding future collusion with the person (gray). The robot verbally clarifies its actions to the non-expert to establish trust. Subsequently, the non-expert directs the robot to a new trajectory using hand gestures. The robot, then, plans a new path passing on the desired side (green). The robot verbally explains itself again while routing on the new path.}
    \label{fig:intro}
    \vspace*{-5mm}
\end{figure}

This study proposes a bidirectional audio-visual human-robot interaction framework for social robot navigation. The framework allows robots to share the decision-making process with nearby people and update their actions based on the response from humans. The framework includes a navigation module for mobile robots in a smart factory environment, considering the social dynamics and visual feedback obtained from the people around. Moreover, the robot explains the additional constraints of both conditions verbally to establish long-term trust between humans. Our social navigation module utilizes a graph attention neural network (GAT) \cite{velivckovic2017graph} backbone to extract social dynamics and track the global context of the environment, enabling accurate trajectory estimation. The route is conditioned both on the graph representation of forecasted trajectories and visual feedback received from nearby humans, with our trustworthy AI module explicitly correlating the decision made based on these conditions.

In summary, this paper introduces a novel end-to-end pipeline to foster trustworthy human-robot interaction for social navigation in collaborative environments, as establishing a communication channel between the robot and the human strengthens trust \cite{che2020efficient}. Our contributions can be outlined as follows:
\begin{itemize}
    \item  A Graph Attention Neural Network-based architecture for social navigation, capable of forecasting trajectories based on the context among individuals.
    \item A trustworthy AI module that rationalizes decisions made, taking into account visual feedback from humans and predicted trajectories
    \item Bidirectional human-robot interaction, recognizing human feedback through hand gestures and providing vocal responses to explain the robot's decision-making process.
\end{itemize}

We initially classified hand gestures for our still-under-development experiments and utilized the Trajnet++ \cite{Kothari2020HumanTF} dataset to extract trajectory features. We plan to design a survey for non-experts to evaluate the proposed framework and to validate our approach further in a smart factory environment using industrial mobile robots once it is established.   


\section{Related Work}

\subsection{Social Navigation for Mobile Robots}
\tuba{Social robot navigation has been widely studied to ensure human-like crowd-aware navigation for safety and reliance \cite{charalampous2017recent}, as it is closely related to forecasting since we expect robots to follow similar social norms as people do. Alahi \& Goel et al. \cite{alahi2016social} exploited LSTM layers to encode trajectories and related nearby people by a mechanism called social pooling. Oh et al. \cite{10160270} combined the social force model \cite{helbing1995social} and Boids \cite{reynolds1987flocks} with the Y-net \cite{mangalam2021goals} to generate realistic pedestrian trajectories. Monte Carlo Tree Search is executed to estimate the costs of candidate goals considering the trajectories of pedestrians. However, the assumptions about the trajectory models oversimplify the human motion. }

\tuba{Some approaches employ deep reinforcement learning techniques to improve social navigation \cite{10160715}. Narayanan et al. \cite{10161504} introduced a transformer network wherein an attention mechanism keeps track of the emotion context to predict trajectories. The prediction and state information are utilized to train a policy network. Chen et al. \cite{chen2017decentralized} developed CADRL, a deep reinforcement learning network, for multi-agent collision avoidance, which utilizes a two-agent value network. However, deep reinforcement learning techniques necessitate reward engineering \cite{chen2017socially}, which is not readily generalizable for real-world applications. Yildirim and Ugur \cite{yildirimlearning} proposed to utilize Conditional Neural Processes Networks to adopt both global and local trajectories, thereby eliminating the necessity for manually designed reward functions. }
%
%


\tuba{There are approaches exploiting graph neural networks to enhance trajectory prediction and social navigation \cite{mo2022multi}, \cite{yang2022ptpgc}, \cite{tang2022evostgat}, \cite{da2022path}. Liu et al. \cite{10115223} utilize GNNs to learn object, obstacle, and robot interactions as relational graphs as the input state for robot policy learning. Similarly, \cite{zhou2024learning} and \cite{escudie2024} leverage GATs to learn state representations as graphs fed into the policy networks. }

\subsection{Trustworthiness in Robotics}

\tuba{Trustworthy AI \cite{dovsilovic2018explainable} has garnered widespread attention as human-robot collaboration becomes increasingly prevalent in today's society \cite{bauer2008human}, \cite{kirtay2021modeling}, \cite{kirtay2022trustworthiness}. Brandao et al. \cite{brandao2021towards} suggested searching planning solutions for various initial states, goals, and trajectories, addressing questions such as ``Why does it fail?" and ``Why trajectory A instead of B?" However, employing brute-force search algorithms is impractical in wide and complex planning domains due to inefficiency. Kottinger et al. \cite{kottinger2021maps} focused on explaining multi-agent motion planning by visualizing and segmenting planned trajectories. Diehl and Ramirez-Amaro \cite{diehl2023causal} utilized Bayesian networks to elaborate failures in cube stacking tasks. Luebbers et al. \cite{luebbers2023autonomous} designed a system in which a drone suggests paths to its human collaborator to justify trajectories. Das et al. \cite{das2021explainable} proposed an encoder-decoder network to generate explanations for errors in pick-and-place tasks, considering the planning context and history. 
}
\begin{figure}
    
    \centering
    \includegraphics[ width=0.5\textwidth]{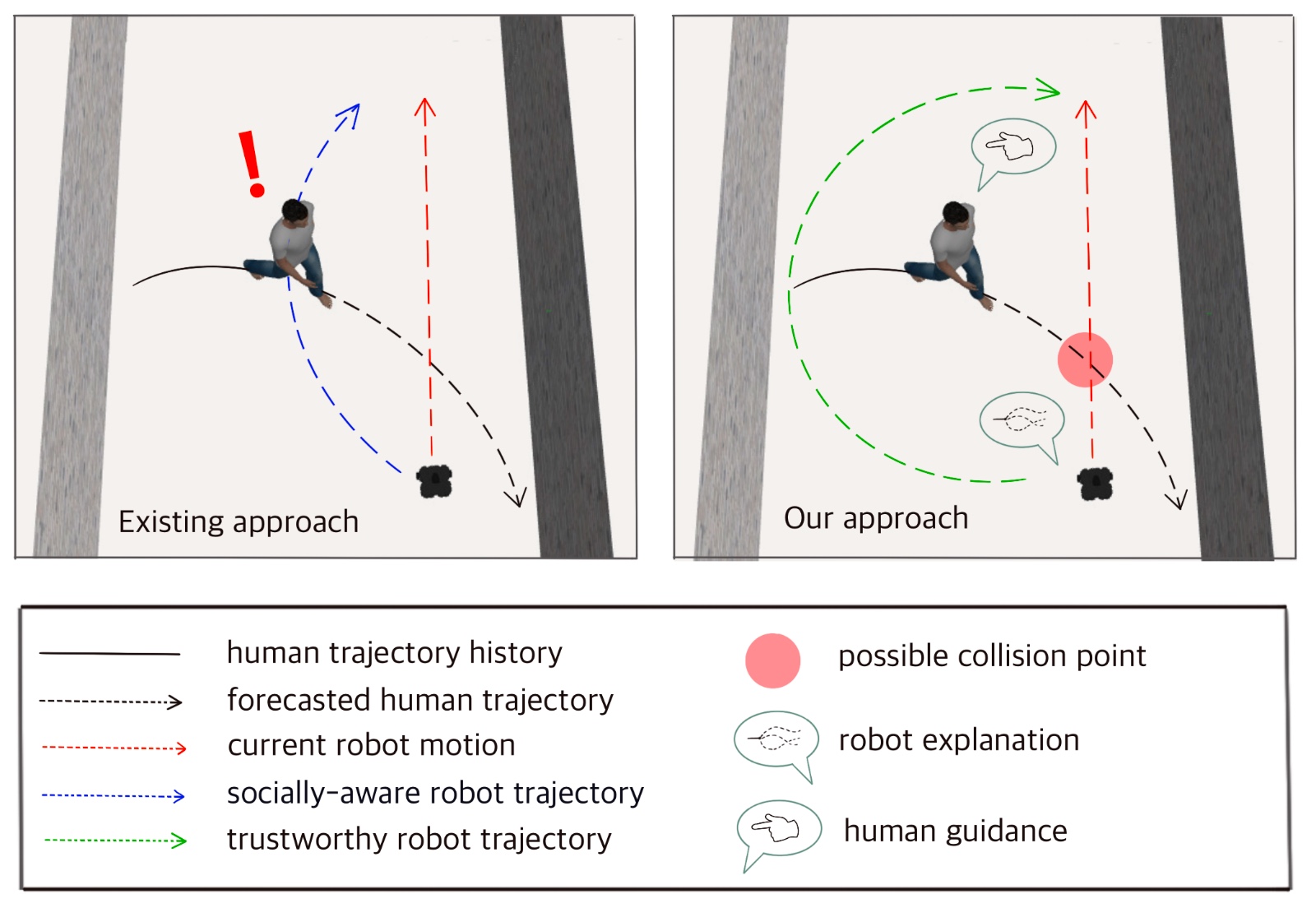}
    \caption{A potential scenario of social robot navigation is depicted. The robot adjusts its original path, taking into account the predicted trajectory of an encountered human. However, the robot initiates movement toward the current position of the human, which may induce fear and erode trust. In the second figure, the proposed system is illustrated. To address the discomfort caused by the robot's unpredictable movements, the proposed system explains the trajectory, expecting guidance in return. 
 }
 \label{fig:bi-directional-communication}
 \vspace*{-5mm}
\end{figure}

As noted in \cite{singamaneni2024survey}, emitting such social cues is highly beneficial to decrease discomfort, which in turn would promote trust. In contrast to prior research, our approach integrates bidirectional audio and visual interactions into social robot navigation by providing explanations for potential paths to receive guidance from humans.

\section{Method}

This study proposes a framework that simultaneously detects and localizes an arbitrary number of people through sensor fusion and predicts their trajectory using LSTM encoders and graph attention networks. Moreover, in case of conflict between the robot's path and a person's predicted trajectory, the robot dynamically adjusts its path and communicates with the person to explain its behavior. The robot is able to readjust its path depending on the visual signal collected from the human subject, which leads to a safer path where cost optimality is not guaranteed. Fig. \ref{fig:bi-directional-communication} shows an example scenario where the trivial path intersects with the predicted human trajectory, and the robot plans an alternative path through the human's current position that is collision-free until the robot reaches that point. However, since this behavior may be uncomfortable for the human, the human redirects the mobile robot by hand gestures. The robot vocally describes the planned path to the human in both stages. After receiving the signal from the human, the robot demonstrates a wider angle maneuver around the person to provide more trustworthy social navigation in a human-oriented environment.

\begin{figure}
    \centering
    \includegraphics[ width=0.5\textwidth]{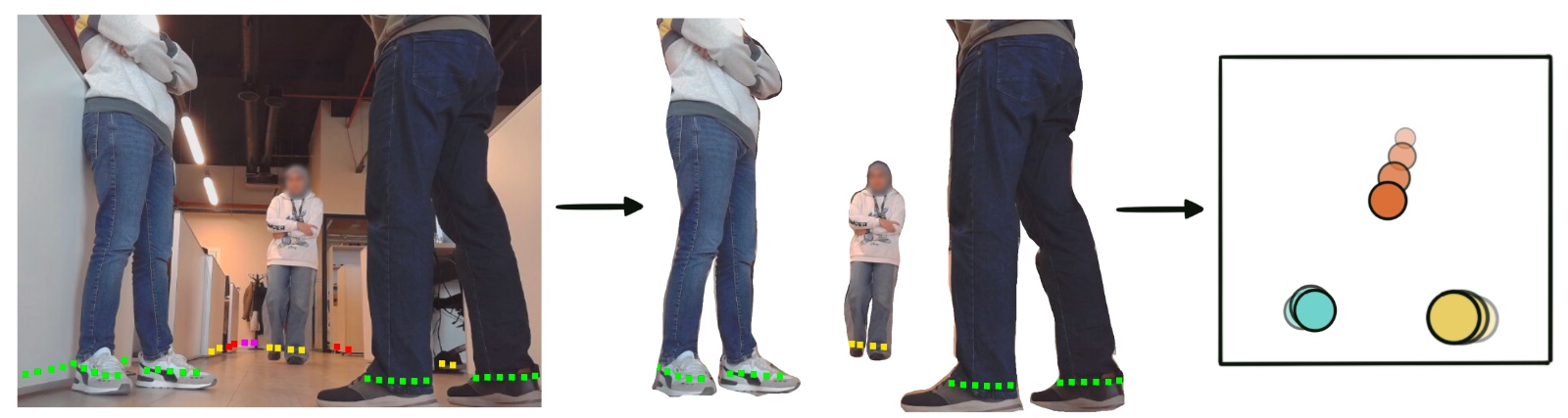}
    \caption{An example of a human detection and localization pipeline in our office environment. The point cloud (colored by distance) is represented on the image plane by weak perspective projection and used for 3D localization by fusing with an instance segmentation algorithm. The 3D estimation locations are converted to trajectories by optical flow between consecutive frames.}
    \label{fig:human_localization}
    \vspace*{-5mm}
\end{figure}

We localize people in the image by fusing an RGB image and a point cloud generated by a 2D LIDAR to provide a cost-effective solution for industrial-grade performance. After projecting the point cloud onto the image plane using weak perspective projection \cite{szeliski2022computer}, we run an instance segmentation algorithm \cite{bolya2019yolact} to find people on 2D images. The average of the set of points falling on the segmentation mask approximates the position of that person, and we convert the positions to trajectories ($T_i$) by using optical flow \cite{szeliski2022computer} in between consecutive frames. We convert the predicted positions into an occupancy grid map to avoid human-robot collisions. Fig. \ref{fig:human_localization} demonstrates an example human localization pipeline on a custom sample. 

\begin{figure}[t]
    \centering
    \includegraphics[ width=0.5\textwidth]{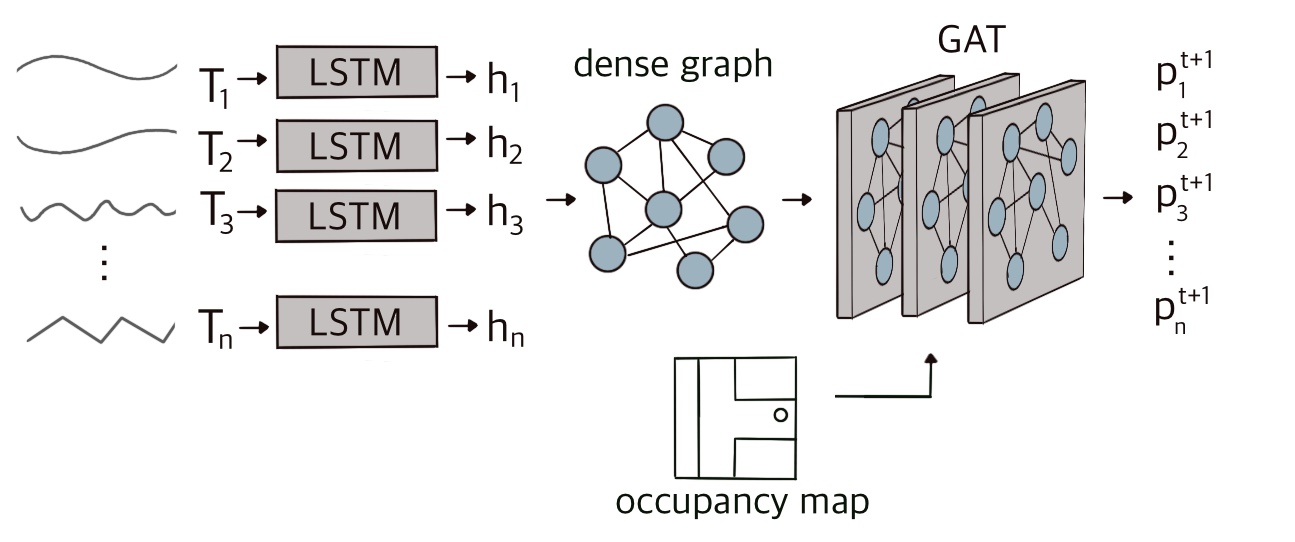}
    \vspace*{-2.5mm}
    \caption{Proposed GAT-based trajectory estimation framework. Human trajectories are encoded with a pretrained LSTM encoder. A dense graph is formed by encodings to track the relation and context between human trajectories using GAT layers. Also, an occupancy map is utilized to static obstacles in the scene.}
    \label{fig:system_overview}
    \vspace*{-5mm}
\end{figure}

The generated trajectories ($T_i$) are fed to the trajectory prediction module to predict the future position of the tracked person. Each of the trajectories is encoded ($h_i$) by a pre-trained LSTM network \cite{sak2014long}. We use the same network to handle an arbitrary number of people in the scene. The encoded trajectories are transformed into a dense graph to not force any relation between trajectories manually. A graph attention network (GAT) predicts the next position ($p^{t+1}_i$) (Fig. \ref{fig:system_overview}). The GAT dynamically learns the correlation between people's trajectories by exploiting both relative position, history, and occupancy map. Since graph neural networks can process graphs of different sizes, the GAT is also capable of processing arbitrary number of people and allows us to make predictions for all people in the scene simultaneously, rather than performing inference for each person separately. We run the pipeline iteratively, concatenating the predicted position ($p^{t+1}_i$) at the end of the current trajectory ($T_i$) to further predict future positions.     

The robot considers predicted future trajectories as obstacles, but not historical positions. As a result, it may plan a path that crosses through the current or nearby position of people and moves towards them. This behavior may cause discomfort for the person, so we designed a trustworthy AI module that plans a new path based on the visual feedback received from the human respondent.

The perception system actively searches for hand gesture signals to adaptively adjust its behavior and respond to the human subject verbally. Hand gestures are classified \cite{lugaresi2019mediapipe} into five groups: `wait', `go left', `go right', `continue', and `unknown' as depicted in Fig. \ref{fig:hand-gestures}. For details see Section \ref{sec:experiments}. To redirect the robot to either \textit{left} or \textit{right}, artificial obstacles are introduced around the person's current position, tilted towards the opposite of the desired direction. To make the robot \textit{wait}, we abort the current mission until the person moves away from the intersection. Similarly, we simply ignore the humans and do not add them to the occupation map as an obstacle to \textit{continue}.

\begin{table}[h]
\caption{List of Vocal Explainations}
\vspace*{-2.5mm}
\label{tab:candidate-sentences}
\begin{center}
\begin{tabular}{l}
\hline
\multicolumn{1}{|l|}{I'm going straight to avoid future collusion.} \\ \hline
\multicolumn{1}{|l|}{I'm passing on the right to avoid future collusion.} \\ \hline
\multicolumn{1}{|l|}{I'm passing on the left to avoid future collusion.} \\ \hline
\multicolumn{1}{|l|}{I'm passing on the right as desired.}                           \\ \hline
\multicolumn{1}{|l|}{I'm passing on the left as desired.}                            \\ \hline
\multicolumn{1}{|l|}{I continue my route, as desired.}                                      \\ \hline
\multicolumn{1}{|l|}{I wait until you pass, as desired.}                                      \\ \hline
\end{tabular}
\end{center}
\vspace*{-2.5mm}
\end{table}

\begin{figure*}[t]
    \centering
    \includegraphics[ width=16cm]{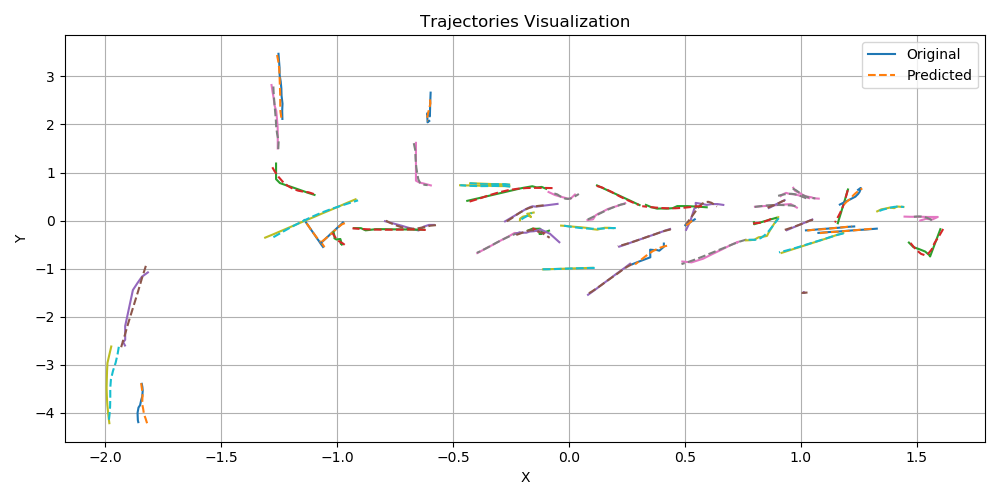}
    \vspace*{-5mm}
    \caption{Self-supervised trajectory encoder output on Trajnet++ dataset. An encoder\&decoder architecture is trained to use the encoder part for GAT input. The architecture successfully recovers the encoded trajectories from the test set.}
    \label{fig:50traj}
    \vspace*{-5mm}
\end{figure*}

For each artificial obstacle layer introduced to the occupancy grid map, we plan a new route \cite{quigley2009ros} that avoids that obstacle. Therefore, we pair each artificial occlusion with a previous and a new path to compare the displacement between the planned position and the one the robot would have if that obstacle were not present. This correspondence allows the robot to explain its behavior to the human, justify the new path and location, and enhance trust between the human and the robot. Once a signal for explanation is received, the robot will state one of the verbal explanations listed in Table \ref{tab:candidate-sentences}.

In order to compare the current position and the one the robot would have if that obstacle were not present we simply check the angle between the actual moving direction and the surrogate direction starting from a common position. If the angle is positive, the moving direction is to the left, and right otherwise. 

\section{Preliminary Experiments} \label{sec:experiments}


The proposed system has localization, tracking, trajectory forecasting, social navigation, and gesture detection components. While the entire navigation system is not yet complete, we have initiated experimentation with selected components. In the following, we present the preliminary results obtained from these components. Integration of these individuals into a full system will be addressed in subsequent stages of our research.

\begin{figure}[h]
    \centering
    \includegraphics[ width=0.45\textwidth]{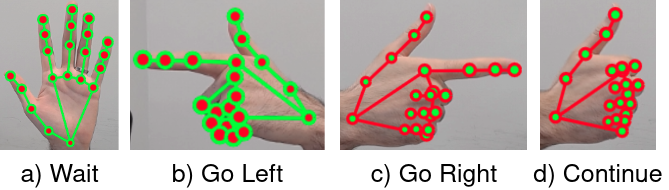}
    \vspace*{-2.5mm}
    \caption{Hand gesture references. Each hand gesture captured is assigned one of these classes or non-classified by the MPJAE distance metric. }
    \label{fig:hand-gestures}
    \vspace*{-5mm}
\end{figure}

To localize humans, we transformed the point cloud to the camera frame using the explicit transformation from the LIDAR plane. We then used weak-perspective projection to project the 3D point cloud onto the image plane. For hand-gesture detection and classification, we directly utilized Mediapipe's \cite{lugaresi2019mediapipe} hand landmark detection model. The distance between the predicted and reference poses was used to classify each hand pose. Others that exceeds the distance threshold for any of the references classified as `unknown` and continues to existing navigation. The reference poses are listed in Fig. \ref{fig:hand-gestures}. The Mean Per Joint Angle Error (MPJAE) was used as the distance metric.

To estimate trajectories, we applied our encoding approach to the Trajnet++ dataset. Fig. \ref{fig:50traj} shows an example of the visual output of the LSTM-based autoencoder on the test set. The LSTM encoder effectively represents the trajectories in the latent space and can reconstruct them. It is important to note that each trajectory will be encoded to the latent space and then used as the nodes of the graph that GAT will process. As performance metrics, we will use Average Displacement Error (ADE) between each position step, Final Destination Error (FDE) considering only the last position accuracy, and Relative Mean Error (RME) to eliminate the error accumulation. As a distance metric, we will use Euclidean distance ($L_2$). To evaluate the social navigation module of our system, we will compare it with CADRL \cite{chen2017decentralized} as a baseline, considering the aforementioned metrics.




\section{Conclusion and Future Works}

At this stage of the study, we have proposed a trustworthy social navigation framework with bidirectional communication. Additionally, we have implemented the preliminary sections for human localization, trajectory encoding, and hand gesture classification. In the future, we plan to finalize the integration of the components and validate our proposal using real-world data collected in a relevant environment. We plan to collect a custom dataset in a smart factory environment where humans and autonomous mobile robots (AMRs) work collaboratively. We will also verify our dataset on the proposed GAT architecture for trajectory forecasting. 


To assess the AI module's trustworthiness, we plan to survey individuals in smart factory settings regarding their comfort with the robot's vocal communication and the effectiveness and variety of hand gestures for controlling the robot. Additionally, as part of the ablation study, our system will be compared in the survey to a version where the human does not redirect the robot at all. Participants will rate their comfort and satisfaction with the robot's vocal responses using a list of candidate sentences. 
Furthermore, participants will redirect the robot and offer feedback on the ease of navigating it using hand gestures in various scenarios.

\bibliographystyle{IEEEtran}
\bibliography{references}

\begin{thebibliography}{10}
\providecommand{\url}[1]{#1}
\csname url@samestyle\endcsname
\providecommand{\newblock}{\relax}
\providecommand{\bibinfo}[2]{#2}
\providecommand{\BIBentrySTDinterwordspacing}{\spaceskip=0pt\relax}
\providecommand{\BIBentryALTinterwordstretchfactor}{4}
\providecommand{\BIBentryALTinterwordspacing}{\spaceskip=\fontdimen2\font plus
\BIBentryALTinterwordstretchfactor\fontdimen3\font minus \fontdimen4\font\relax}
\providecommand{\BIBforeignlanguage}[2]{{%
\expandafter\ifx\csname l@#1\endcsname\relax
\typeout{** WARNING: IEEEtran.bst: No hyphenation pattern has been}%
\typeout{** loaded for the language `#1'. Using the pattern for}%
\typeout{** the default language instead.}%
\else
\language=\csname l@#1\endcsname
\fi
#2}}
\providecommand{\BIBdecl}{\relax}
\BIBdecl

\bibitem{sheridan2016human}
T.~B. Sheridan, ``Human--robot interaction: status and challenges,'' \emph{Human factors}, vol.~58, no.~4, pp. 525--532, 2016.

\bibitem{bartneck2020human}
C.~Bartneck, T.~Belpaeme, F.~Eyssel, T.~Kanda, M.~Keijsers, and S.~{\v{S}}abanovi{\'c}, \emph{Human-robot interaction: An introduction}.\hskip 1em plus 0.5em minus 0.4em\relax Cambridge University Press, 2020.

\bibitem{samek2021explaining}
W.~Samek, G.~Montavon, S.~Lapuschkin, C.~J. Anders, and K.-R. M{\"u}ller, ``Explaining deep neural networks and beyond: A review of methods and applications,'' \emph{Proceedings of the IEEE}, vol. 109, no.~3, pp. 247--278, 2021.

\bibitem{hancock2023and}
P.~Hancock, T.~T. Kessler, A.~D. Kaplan, K.~Stowers, J.~C. Brill, D.~R. Billings, K.~E. Schaefer, and J.~L. Szalma, ``How and why humans trust: A meta-analysis and elaborated model,'' \emph{Frontiers in psychology}, vol.~14, 2023.

\bibitem{brandao2021towards}
M.~Brandao, G.~Canal, S.~Krivi{\'c}, and D.~Magazzeni, ``Towards providing explanations for robot motion planning,'' in \emph{2021 IEEE International Conference on Robotics and Automation (ICRA)}.\hskip 1em plus 0.5em minus 0.4em\relax IEEE, 2021, pp. 3927--3933.

\bibitem{fox2017explainable}
M.~Fox, D.~Long, and D.~Magazzeni, ``Explainable planning,'' \emph{arXiv preprint arXiv:1709.10256}, 2017.

\bibitem{almagor2020explainable}
S.~Almagor and M.~Lahijanian, ``Explainable multi agent path finding,'' in \emph{AAMAS}, 2020.

\bibitem{kottinger2021maps}
J.~Kottinger, S.~Almagor, and M.~Lahijanian, ``Maps-x: Explainable multi-robot motion planning via segmentation,'' in \emph{2021 IEEE International Conference on Robotics and Automation (ICRA)}.\hskip 1em plus 0.5em minus 0.4em\relax IEEE, 2021, pp. 7994--8000.

\bibitem{kottinger2022conflict}
------, ``Conflict-based search for explainable multi-agent path finding,'' in \emph{Proceedings of the International Conference on Automated Planning and Scheduling}, vol.~32, 2022, pp. 692--700.

\bibitem{das2021explainable}
D.~Das, S.~Banerjee, and S.~Chernova, ``Explainable ai for robot failures: Generating explanations that improve user assistance in fault recovery,'' in \emph{Proceedings of the 2021 ACM/IEEE International Conference on Human-Robot Interaction}, 2021, pp. 351--360.

\bibitem{diehl2023causal}
M.~Diehl and K.~Ramirez-Amaro, ``A causal-based approach to explain, predict and prevent failures in robotic tasks,'' \emph{Robotics and Autonomous Systems}, vol. 162, p. 104376, 2023.

\bibitem{angelopoulos2022you}
G.~Angelopoulos, A.~Rossi, C.~Di~Napoli, and S.~Rossi, ``You are in my way: non-verbal social cues for legible robot navigation behaviors,'' in \emph{2022 IEEE/RSJ International Conference on Intelligent Robots and Systems (IROS)}.\hskip 1em plus 0.5em minus 0.4em\relax IEEE, 2022, pp. 657--662.

\bibitem{mavrogiannis2019effects}
C.~Mavrogiannis, A.~M. Hutchinson, J.~Macdonald, P.~Alves-Oliveira, and R.~A. Knepper, ``Effects of distinct robot navigation strategies on human behavior in a crowded environment,'' in \emph{2019 14th ACM/IEEE International Conference on Human-Robot Interaction (HRI)}.\hskip 1em plus 0.5em minus 0.4em\relax IEEE, 2019, pp. 421--430.

\bibitem{velivckovic2017graph}
P.~Veli{\v{c}}kovi{\'c}, G.~Cucurull, A.~Casanova, A.~Romero, P.~Lio, and Y.~Bengio, ``Graph attention networks,'' \emph{arXiv preprint arXiv:1710.10903}, 2017.

\bibitem{che2020efficient}
Y.~Che, A.~M. Okamura, and D.~Sadigh, ``Efficient and trustworthy social navigation via explicit and implicit robot--human communication,'' \emph{IEEE Transactions on Robotics}, vol.~36, no.~3, pp. 692--707, 2020.

\bibitem{Kothari2020HumanTF}
P.~Kothari, S.~Kreiss, and A.~Alahi, ``Human trajectory forecasting in crowds: A deep learning perspective,'' \emph{IEEE Transactions on Intelligent Transportation Systems}, pp. 1--15, 2021.

\bibitem{charalampous2017recent}
K.~Charalampous, I.~Kostavelis, and A.~Gasteratos, ``Recent trends in social aware robot navigation: A survey,'' \emph{Robotics and Autonomous Systems}, vol.~93, pp. 85--104, 2017.

\bibitem{alahi2016social}
A.~Alahi, K.~Goel, V.~Ramanathan, A.~Robicquet, L.~Fei-Fei, and S.~Savarese, ``Social lstm: Human trajectory prediction in crowded spaces,'' in \emph{Proceedings of the IEEE conference on computer vision and pattern recognition}, 2016, pp. 961--971.

\bibitem{10160270}
J.~Oh, J.~Heo, J.~Lee, G.~Lee, M.~Kang, J.~Park, and S.~Oh, ``Scan: Socially-aware navigation using monte carlo tree search,'' in \emph{2023 IEEE International Conference on Robotics and Automation (ICRA)}, 2023, pp. 7576--7582.

\bibitem{helbing1995social}
D.~Helbing and P.~Molnar, ``Social force model for pedestrian dynamics,'' \emph{Physical review E}, vol.~51, no.~5, p. 4282, 1995.

\bibitem{reynolds1987flocks}
C.~W. Reynolds, ``Flocks, herds and schools: A distributed behavioral model,'' in \emph{Proceedings of the 14th annual conference on Computer graphics and interactive techniques}, 1987, pp. 25--34.

\bibitem{mangalam2021goals}
K.~Mangalam, Y.~An, H.~Girase, and J.~Malik, ``From goals, waypoints \& paths to long term human trajectory forecasting,'' in \emph{Proceedings of the IEEE/CVF International Conference on Computer Vision}, 2021, pp. 15\,233--15\,242.

\bibitem{10160715}
Y.-J. Mun, M.~Itkina, S.~Liu, and K.~Driggs-Campbell, ``Occlusion-aware crowd navigation using people as sensors,'' in \emph{2023 IEEE International Conference on Robotics and Automation (ICRA)}, 2023, pp. 12\,031--12\,037.

\bibitem{10161504}
V.~Narayanan, B.~M. Manoghar, R.~P. RV, and A.~Bera, ``Ewarenet: Emotion-aware pedestrian intent prediction and adaptive spatial profile fusion for social robot navigation,'' in \emph{2023 IEEE International Conference on Robotics and Automation (ICRA)}, 2023, pp. 7569--7575.

\bibitem{chen2017decentralized}
Y.~F. Chen, M.~Liu, M.~Everett, and J.~P. How, ``Decentralized non-communicating multiagent collision avoidance with deep reinforcement learning,'' in \emph{2017 IEEE international conference on robotics and automation (ICRA)}.\hskip 1em plus 0.5em minus 0.4em\relax IEEE, 2017, pp. 285--292.

\bibitem{chen2017socially}
Y.~F. Chen, M.~Everett, M.~Liu, and J.~P. How, ``Socially aware motion planning with deep reinforcement learning,'' in \emph{2017 IEEE/RSJ International Conference on Intelligent Robots and Systems (IROS)}.\hskip 1em plus 0.5em minus 0.4em\relax IEEE, 2017, pp. 1343--1350.

\bibitem{yildirimlearning}
Y.~Yildirim and E.~Ugur, ``Learning social navigation from demonstrations with deep neural networks,'' \emph{arXiv preprint arXiv:2404.11246}, 2024.

\bibitem{mo2022multi}
X.~Mo, Z.~Huang, Y.~Xing, and C.~Lv, ``Multi-agent trajectory prediction with heterogeneous edge-enhanced graph attention network,'' \emph{IEEE Transactions on Intelligent Transportation Systems}, vol.~23, no.~7, pp. 9554--9567, 2022.

\bibitem{yang2022ptpgc}
J.~Yang, X.~Sun, R.~G. Wang, and L.~X. Xue, ``Ptpgc: Pedestrian trajectory prediction by graph attention network with convlstm,'' \emph{Robotics and Autonomous Systems}, vol. 148, p. 103931, 2022.

\bibitem{tang2022evostgat}
H.~Tang, P.~Wei, J.~Li, and N.~Zheng, ``Evostgat: Evolving spatiotemporal graph attention networks for pedestrian trajectory prediction,'' \emph{Neurocomputing}, vol. 491, pp. 333--342, 2022.

\bibitem{da2022path}
F.~Da and Y.~Zhang, ``Path-aware graph attention for hd maps in motion prediction,'' in \emph{2022 International Conference on Robotics and Automation (ICRA)}.\hskip 1em plus 0.5em minus 0.4em\relax IEEE, 2022, pp. 6430--6436.

\bibitem{10115223}
Z.~Liu, Y.~Zhai, J.~Li, G.~Wang, Y.~Miao, and H.~Wang, ``Graph relational reinforcement learning for mobile robot navigation in large-scale crowded environments,'' \emph{IEEE Transactions on Intelligent Transportation Systems}, vol.~24, no.~8, pp. 8776--8787, 2023.

\bibitem{zhou2024learning}
Y.~Zhou and J.~Garcke, ``Learning crowd behaviors in navigation with attention-based spatial-temporal graphs,'' \emph{arXiv preprint arXiv:2401.06226}, 2024.

\bibitem{escudie2024}
E.~Escudie and L.~M.~J. Saraydaryan, ``Attention graph for multi-robot social navigation with deep reinforcement learning,'' in \emph{International Conference on Autonomous Agents and Multiagent Systems (AAMAS)}, 2024.

\bibitem{dovsilovic2018explainable}
F.~K. Do{\v{s}}ilovi{\'c}, M.~Br{\v{c}}i{\'c}, and N.~Hlupi{\'c}, ``Explainable artificial intelligence: A survey,'' in \emph{2018 41st International convention on information and communication technology, electronics and microelectronics (MIPRO)}.\hskip 1em plus 0.5em minus 0.4em\relax IEEE, 2018, pp. 0210--0215.

\bibitem{bauer2008human}
A.~Bauer, D.~Wollherr, and M.~Buss, ``Human--robot collaboration: a survey,'' \emph{International Journal of Humanoid Robotics}, vol.~5, no.~01, pp. 47--66, 2008.

\bibitem{kirtay2021modeling}
M.~Kirtay, E.~Oztop, M.~Asada, and V.~V. Hafner, ``Modeling robot trust based on emergent emotion in an interactive task,'' in \emph{2021 IEEE International Conference on Development and Learning (ICDL)}.\hskip 1em plus 0.5em minus 0.4em\relax IEEE, 2021, pp. 1--8.

\bibitem{kirtay2022trustworthiness}
M.~Kirtay, E.~Oztop, A.~K. Kuhlen, M.~Asada, and V.~V. Hafner, ``Trustworthiness assessment in multimodal human-robot interaction based on cognitive load,'' in \emph{2022 31st IEEE International Conference on Robot and Human Interactive Communication (RO-MAN)}.\hskip 1em plus 0.5em minus 0.4em\relax IEEE, 2022, pp. 469--476.

\bibitem{luebbers2023autonomous}
M.~B. Luebbers, A.~Tabrez, K.~Ruvane, and B.~Hayes, ``Autonomous justification for enabling explainable decision support in human-robot teaming,'' \emph{Proceedings of Robotics: Science and Systems. Daegu, Republic of Korea. https://doi. org/10.15607/RSS}, 2023.

\bibitem{singamaneni2024survey}
P.~T. Singamaneni, P.~Bachiller-Burgos, L.~J. Manso, A.~Garrell, A.~Sanfeliu, A.~Spalanzani, and R.~Alami, ``A survey on socially aware robot navigation: Taxonomy and future challenges,'' \emph{The International Journal of Robotics Research}, p. 02783649241230562, 2024.

\bibitem{szeliski2022computer}
R.~Szeliski, \emph{Computer vision: algorithms and applications}.\hskip 1em plus 0.5em minus 0.4em\relax Springer Nature, 2022.

\bibitem{bolya2019yolact}
D.~Bolya, C.~Zhou, F.~Xiao, and Y.~J. Lee, ``Yolact: Real-time instance segmentation,'' in \emph{Proceedings of the IEEE/CVF international conference on computer vision}, 2019, pp. 9157--9166.

\bibitem{sak2014long}
H.~Sak, A.~Senior, and F.~Beaufays, ``Long short-term memory based recurrent neural network architectures for large vocabulary speech recognition,'' \emph{arXiv preprint arXiv:1402.1128}, 2014.

\bibitem{lugaresi2019mediapipe}
C.~Lugaresi, J.~Tang, H.~Nash, C.~McClanahan, E.~Uboweja, M.~Hays, F.~Zhang, C.-L. Chang, M.~G. Yong, J.~Lee \emph{et~al.}, ``Mediapipe: A framework for building perception pipelines,'' \emph{arXiv preprint arXiv:1906.08172}, 2019.

\bibitem{quigley2009ros}
M.~Quigley, K.~Conley, B.~Gerkey, J.~Faust, T.~Foote, J.~Leibs, R.~Wheeler, A.~Y. Ng \emph{et~al.}, ``Ros: an open-source robot operating system,'' in \emph{ICRA workshop on open source software}, vol.~3, no. 3.2.\hskip 1em plus 0.5em minus 0.4em\relax Kobe, Japan, 2009, p.~5.

\end{thebibliography}
\end{document}